\pdfoutput=1 
\documentclass[11pt,a4paper]{article}
\usepackage{coling2020}
\usepackage{times}
\usepackage{latexsym}
\usepackage{graphicx}
\usepackage{amsmath}

\newcommand{\citep}{\cite}
\newcommand{\citet}{\newcite}
\usepackage{xcolor}
\usepackage{multirow}
\usepackage{url}

\colingfinalcopy
\usepackage{microtype}

\title{Semi-supervised URL Segmentation with Recurrent Neural Networks Pre-trained on Knowledge Graph Entities}
\author{Hao Zhang \and Jae Ro \and Richard Sproat \\
        Google Research \\
         \texttt{\{haozhang,jaero,rws\}@google.com}}
\date{}

\begin{document}
\maketitle
\begin{abstract}
Breaking domain names such as \texttt{openresearch} into component words \texttt{open} and \texttt{research} is important for applications like Text-to-Speech synthesis and web search. We link this problem to the classic problem of Chinese word segmentation and show the effectiveness of a tagging model based on Recurrent Neural Networks (RNNs) using characters as input. To compensate for the lack of training data, we propose a pre-training method on concatenated entity names in a large knowledge database. Pre-training improves the model by 33\% and brings the sequence accuracy to 85\%.
\end{abstract}

\section{Introduction}

%
%
\blfootnote{
\hspace{-0.65cm}  
This work is licensed under a Creative Commons Attribution 4.0 International License.
License details:
\url{http://creativecommons.org/licenses/by/4.0/}.
}

Word segmentation is a fundamental NLP analysis problem for written languages with no space delimiters between words such as Chinese and Japanese. 
In the age of digital communications, new URLs (e.g. \texttt{www.openresearch.org}) and hashtags (e.g. \texttt{\#photooftheday}), which often include strings of concatenated words (\texttt{openresearch}, \texttt{photooftheday}) are being added every day to a growing set of tokens that an NLP system may need to deal with, and
they pose challenges for language and speech applications. For example, a Text-to-Speech (TTS) synthesis system will struggle to pronounce these concatenated tokens, since simply applying a grapheme-to-phoneme system out of the box to something like \texttt{photooftheday} will usually yield poor results. This suggests the need for a model that can split such tokens into the component words. 
So-called ``end-to-end'' neural TTS systems \citep{sotelo-etal-2017,Wang-etal-2017-tacotron}, which learn to map directly from character sequences to speech might seem to hold out the hope of avoiding treating this problem separately. However, the fact that URLs occur relatively rarely in most TTS training data limits the promise of such models on this long-tail problem. 

The problem of analyzing URLs does differ in one useful way from more general text normalization problems. For a token such as \texttt{123} in a text, one typically needs to know what context it occurs in in order to know how to read it: is it \texttt{one hundred twenty three} or \texttt{one twenty three}; see \citep{Sproat:EtAl:2001}, inter alia. In the case of URLs, these are largely \emph{context-independent} since the output segmentation is usually unaffected by the surrounding words. Hence the problem can be treated as a standalone one that does not require the system to be trained as part of broader text normalization training.

Our training data comes from 
camel case URLs that naturally define the segment boundaries (e.g. \texttt{NYTimes.com} maps to \texttt{N Y Times . com}) along with manual corrections for non-trivial boundaries. 
We release our training and evaluation data sets to promote research on this problem. 
By drawing an analogy with Chinese word segmentation, we cast the URL segmentation problem as a sequence tagging problem. We propose a simple Recurrent Neural Network (RNN) based tagger with an encoder and a decoder. 

The model trained on the data set has a decent full sequence accuracy (64\%) but fails to generalize to more rare words due to the size of the training data. Inspired by the success of pre-training in many NLP tasks \citep{peters-etal-2017-semi,devlin-etal-2019-bert}, we propose a pre-training recipe for the segmenter. Based on the observation that URLs are often compound entity names and so are knowledge graph entities \citep{bollacker-etal-2008}, we create a large synthetic training data set by concatenating the knowledge graph entity names. We observe 21\% absolute (33\% relative) improvement in sequence accuracy after applying pre-training followed by fine-tuning.

\section{Related Work}
\label{sec:related}

Every text-to-speech system has to be able to read URLs and other electronic addresses, but there is very little in the published literature that discusses this
problem specifically as a research topic, and most systems seem not to do much in the way of interesting analysis of the internal components of the address. For example, the Kestrel text normalization system \citep{ebden-sproat-2014} identifies URLs and other electronic addresses using finite-state matchers, and parses the main components into separate tokens based on standard delimiters (\texttt{/}, \texttt{:}, etc.): thus \texttt{www.google.com} would be parsed into \texttt{www . google . com}. The individual components are then pronounced separately. Common components such as \texttt{.}, \texttt{www}, \texttt{nytimes} and \texttt{com} are handled by table look-up, but other components, such as \texttt{jpopasia} in \texttt{jpopasia.com} are not otherwise broken down and if they do not match a lexicon entry, may end up being read letter-by-letter.

The problem has applications beyond TTS. In web search, analyzing URLs and hashtags leads to better matching.
\citet{wang-etal-2011-webnlp} termed the problem ``URL word breaking''. They used a noisy channel model with an $n$-gram language model trained on word-segmented data and a word-synchronous beam search algorithm for inference. The model is essentially unsupervised. They found that the style of the text used to build the model played a crucial role and document titles yielded the best results. 
In our pre-training experiments, we also tried web queries and documents. None of them gave the same improvement as the knowledge graph entity names in title case.
\citet{srinivasan-etal-2012-hastags} improved the model of \citet{wang-etal-2011-webnlp} by adding a supervised max-margin structured prediction model using individual unsupervised language models as features. Fine-tuning on in-domain data is the counterpart in our system.
Both have created training or evaluation data sets of URL domain names for their experiments, but these have not been publicly released.
We contribute a data set of URLs crawled from a public repository of Web texts with their internal segments annotated by crowd sourcing.

\citet{chiang-etal-2010-bayesian} reported experiments on a related artificial problem of splitting of space-free English through Bayesian inference for FSTs following the work of \citet{goldwater-etal-2009-cognition}.\footnote{It is worth noting that the problem of resegmenting space-free English has a long history: the earliest reference we are aware of is \citep{Olivier:68}.}

A closely related problem is compound splitting. \citet{macherey-etal-2011-language} presented an unsupervised  probabilistic model for splitting compound words into parts, with the compound part sub-model being a zero-order model to enable efficient dynamic programming inference. The model is optimized for the task of machine translation. They only reported results for seven (Germanic, Hellenic, and Uralic) languages other than English. More recently, fully supervised letter sequence labelling models have been introduced for German compound splitting \citep{ma-etal-2016-letter} and Sanskrit word splitting \citep{hellwig-nehrdich-2018-sanskrit}. Pre-training can potentially further improve these models.

There is a large body of research on Chinese word segmentation. The best models are supervised ones using structured prediction \citep{peng-etal-2004-chinese}, transition-based models \citep{zhang-clark-2007-chinese}, and most recently RNNs \citep{ma-etal-2018-state} and BERT-based models \citep{huang-etal-2019-bert-segmentation}. The superior results of the BERT-based models demonstrate that pre-training is effective on word segmentation. Unlike BERT, we pre-train the entire model, not just the encoder.

\section{RNN Tagging Model}
\label{sec:tagging}

\begin{figure}
  \includegraphics[trim=2.25in 1.75in 2in 1.75in,clip=true,width=0.8\columnwidth]{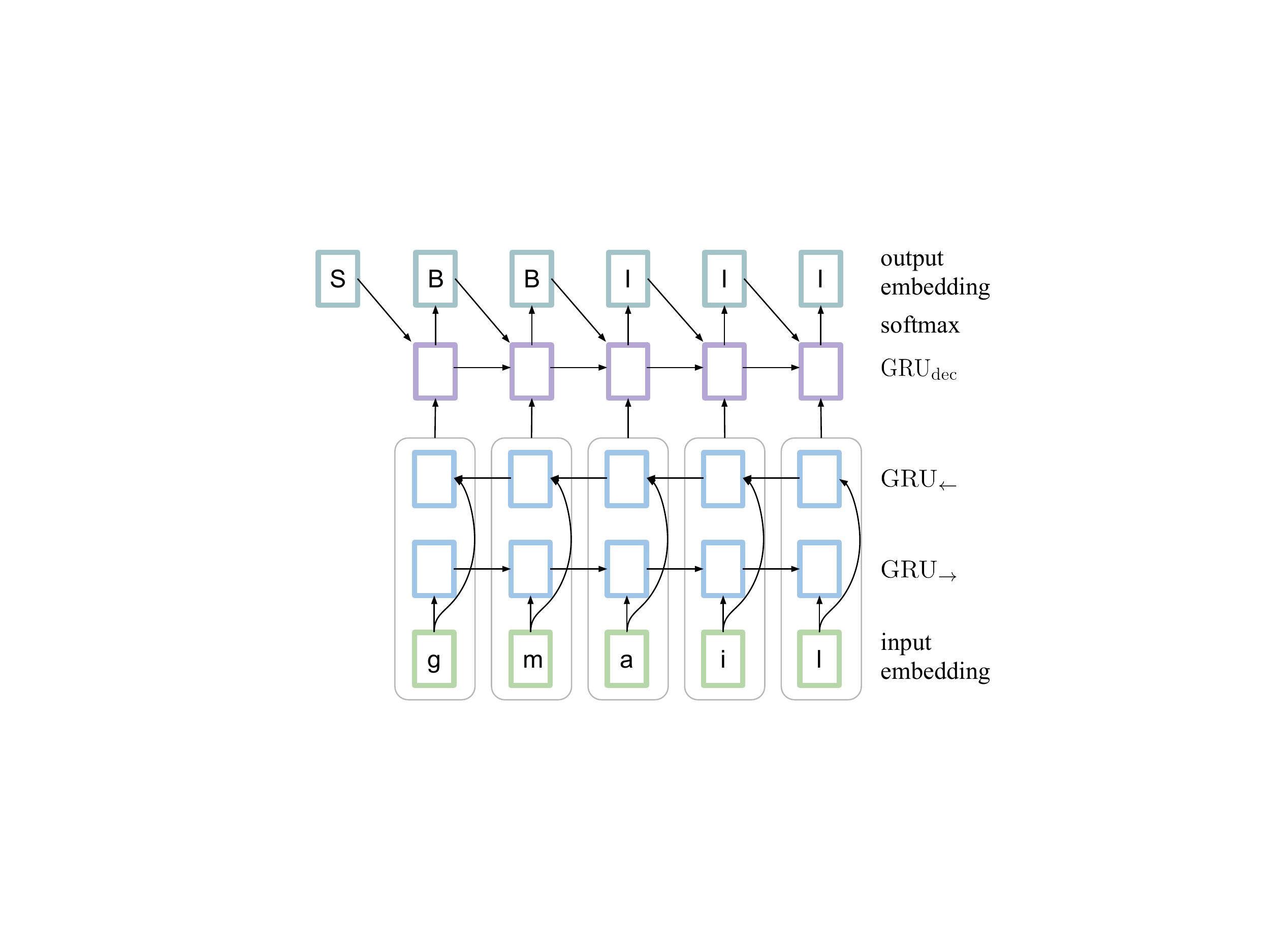}
  \caption{RNN tagging model architecture. $\text{GRU}_{\rightarrow}$ and $\text{GRU}_{\leftarrow}$ are forward and backward encoder RNNs implemented with Gated Recurrent Units (GRU). $\text{GRU}_{dec}$ is the decoder RNN. The concatenation of $\text{GRU}_{\rightarrow}$, $\text{GRU}_{\leftarrow}$, and the input embedding at the current position, as well as the embedding of the previous tag is fed to $\text{GRU}_{dec}$. The output of $\text{GRU}_{dec}$ is fed to a $\text{softmax}$ layer to generate the output tagging sequence.
  }
  \label{fig:rnn_tagger}
\end{figure}

We formulate the segmentation problem as a character sequence tagging model. Given an input character sequence $X=x_1,\dots,x_I$, the model predicts an output tag sequence $Y=y_1, \dots, y_I$, with $y_i \in \{\text{B, I}\}$ being the tag for the character $x_i$. $\text{B}$ indicates the underlying character starts a new segment. $\text{I}$ indicates the underlying character continues the current segment. Tag sequences $\{\text{B, I}\}^+$ correspond one-to-one to segment sequences. For example, $\text{B,B,I,B}$ is equivalent to the segment sequence $[x_1], [x_2, x_3], [x_4]$. 

We model $P(Y|X)$ with an encoder-decoder RNN. The encoder generates hidden state sequences of length $I$. At decoding step $i$, the decoder attends to position $i$ in both the hidden state sequences and the input embedding sequence. 
Figure~\ref{fig:rnn_tagger} illustrates the architecture of the model. Our architecture is a modification of the stacking LSTM architecture of \citet{ma-etal-2018-state} which can also be understood as using hard attention \citep{aharoni-goldberg-2017-morphological} in the decoder.
We do not use bigram character features. Instead, we rely on forward and backward RNNs for implicit input feature extraction. We make the decoder auto-regressive by feeding the previously predicted tag to the decoder RNN and applying beam search in inference.

\subsection{Pre-training}
RNN models demand a large number of training examples. While labelled data is often scarce, un-labelled data with matching domain is often abundant. Pre-training the encoder component of a sequence-to-sequence model for a different task such as language modelling has been extremely successful, as manifested by the BERT model. We take a different approach because we not only can find data that matches the domain of interest but also can construct input-output mappings with high accuracy. Therefore, we pre-train the entire model with the same objective on a synthetic domain-matching data set. The fine-tuning phase follows the pre-training phase by simply switching the training data to the labelled data set.
\section{Data}
\label{sec:data}
\subsection{Camel Case URLs}
We use two distinct sources for our camel case URL data set, an internal crawl and Common Crawl\footnote{https://commoncrawl.org}. The internal data is comprised of approximately 21k domain names automatically segmented based on case (\texttt{DisneylandNews}$\rightarrow$\texttt{Disneyland News}) and then manually corrected. Manually correcting the data created interesting examples where the segmentation, given casing, is non-trivial (e.g. \texttt{Awardsand Honors} vs. \texttt{Awards and Honors}). The Common Crawl data was scraped from extracted plain text from web archives and consists of approximately 21k domain names with manual corrections done via crowd sourcing. The Common Crawl URL data set is publicly available on GitHub.\footnote{https://github.com/google-research-datasets/common-crawl-domain-names}

\subsection{Knowledge Graph Entity Names}
The pre-training data is derived from entity names found in Google's Knowledge Graph\footnote{https://developers.google.com/knowledge-graph}. The entity names are naturally space-separated and case-sensitive. Further splitting is also done on entity name segments that are all uppercase consonants since these are virtually always verbalized letter by letter (e.g. \texttt{CD Player} to \texttt{C D Player}).

\subsection{Data Set Statistics}
Table~\ref{tab:data_stats} lists the statistics of the data sets.
The two data sets, namely the internal crawl and the Common Crawl, have approximately equal means of input and output lengths, even though their creation processes are different. On the other hand, the {\em pre-train} Knowledge Graph data set contains larger proportions of longer sequences. The {\em pre-train} set is four orders of magnitude larger than either of the two annotated data sets.

\begin{table}[htb]
    \centering
    \begin{tabular}{l|c|c|c}
         &  {\em Average } & {\em Average } & {\em Total } \\
         &  {\em Input } & {\em No. of } & {\em No. of} \\
         &  {\em Length } & {\em Segments } & {\em Examples} \\
         \hline
         internal crawl {\em train} & 12.96 & 2.47 & 17036 \\
         internal crawl {\em dev} & 12.94 & 2.48 & 1893 \\
         internal crawl  {\em test} & 13.17 & 2.49 & 2104 \\
         \hline
         Common Crawl {\em train} & 12.63 &	2.65 & 17575 \\
         Common Crawl {\em dev} & 12.77	& 2.66	& 1953 \\
         Common Crawl  {\em test} & 12.64 &	2.67 & 2170 \\
         \hline
         Knowlege Graph {\em pre-train} & 29.22 & 4.54 & $>$200m \\
    \end{tabular}
    \caption{Statistics for camel case URL data sets and Knowledge Graph pre-train data set.}
    \label{tab:data_stats}
\end{table}
\section{Experiments}
We choose the hyper-parameters on the internal crawl dev set for the network in Figure~\ref{fig:rnn_tagger}. Table~\ref{tab:network_specs} lists the key parameters. In Section~\ref{sec:lowercase_training} and Section~\ref{sec:mixedcase_training}, we develop training strategies using the internal crawl data set. In Section~\ref{sec:final_results}, we report the final results on both the internal and the Common Crawl test sets based on the hyper-parameters and the training strategy developed on the internal data set.
\begin{table}[ht]
    \centering
    \begin{tabular}{lr}
             \hline
             Input embedding size & 256 \\
             Output embedding size & 64 \\
             Number of forward encoder layers & 2 \\
             Number of backward encoder layers & 2 \\
             Number of decoder layers & 1 \\
             Number of decoder GRU units & 64 \\
             Number of encoder GRU units per layer & 256\\
             Beam size & 2 \\
             \hline
    \end{tabular}
    \caption{Network hyper-parameters.}
    \label{tab:network_specs}
\end{table}

We report results in terms of full sequence accuracy, where
$\it{Accuracy} = \frac{\it{\# Correct Segmentations}}{\it{\# Sequences}}$.
For example, \texttt{photooftheday} has only one correct segmentation \texttt{photo of the day}. There is no partial credit.

\subsection{Lowercase Training}
\label{sec:lowercase_training}
\begin{table}[ht]
    \centering
    \begin{tabular}{l|c}
             & {\em Lowercase Accuracy} \\
             \hline
    baseline & 70.10\%  \\
    pre-train & 82.25\%  \\
    +fine-tune & {\em 88.80\%}  \\
    \end{tabular}
    \caption{Lowercase results. Training and pre-training data sets are lower-cased. Results are on the development set which is also in lowercase.}
    \label{tab:lowercase_expt}
\end{table}

The camel case training data has implicit word boundary annotations. In order to train a model to predict boundaries when case cues are not present, we need to hide the annotations by normalizing case.  For this we simply lowercase both the training and evaluation sets. Table~\ref{tab:lowercase_expt} summarizes the results of pure lowercase models. The baseline model uses no external data besides the {\em train} set.


The improved model is first trained only on the {\em pre-train} set
and then fine-tuned on the {\em train} set.
Pre-training alone is already better than the baseline by a large margin (12\%), indicating the importance of learning from a large number of entity names. Fine-tuning yields a further improvement (6\%).



\subsection{Mixed Case Training}
\label{sec:mixedcase_training}
\begin{table}[ht]
    \centering
    \begin{tabular}{l|c|c}
         &  {\em Lowercase Accuracy } & {\em Camel Case Accuracy} \\
         \hline
         baseline & 69.36\%  & 94.82\%  \\
         pre-train & 81.56\%  & 92.18\%  \\
         +fine-tune & {\em 89.54\%}  & {\em 96.67\%}  \\
    \end{tabular}
    \caption{Mixed LowerCase/CamelCase results. Training and pre-training sets have equal proportions of lowercase and camel case examples. Results are reported on both lowercase and camel case development sets.}
    \label{tab:mixed_case_expt}
\end{table}
The lowercase model works well on lowercase input. But the accuracy of 88.8\% is not very high for camel case input because 
the simple rule of splitting words based on case switching is 92.55\% accurate.
One could have a hybrid system in which the neural segmenter is invoked when the input has no case cues and the rule is invoked for camel case input, but we ought in principle to be able to train a model that handles both kinds of input well. 

In Table~\ref{tab:mixed_case_expt}, we show our final results of training on mixed lowercase and camel case data. For every camel case example, a lowercase example is added. This is done for both the training set and the pre-training set. Essentially, we assign equal weights to the two types of examples. There is a small degradation in lowercase accuracy for the baseline and the pre-train-only models. But after fine-tuning, not only is the loss recovered, but there is even a slight gain (89.54\% versus 88.80\%). This can be explained as an effect of transfer-learning if we view the lowercase examples and camel case examples as two different domains. As expected, the camel case accuracy is now close to perfect (96.67\%), higher than the accuracy of the simple rule of splitting-by-case (92.55\%). 


\subsection{Final Results}
\label{sec:final_results}
In this section, we report the final results on the {\em test} portion of the internal crawl and the Common Crawl data sets.
The model uses the mixed case training strategy mentioned in Section~\ref{sec:mixedcase_training}. Learning rates are tuned on their {\em dev} counterparts.
The top portion of Table~\ref{tab:crawl_test} is for internal crawl. The bottom portion is for Common Crawl. The common trends are:
\begin{itemize}
    \item Pre-training (without fine-tuning) improves over baseline by a large margin when input is lowercase.
    \item Fine-tuning further improves the results regardless of input casing.
\end{itemize}
\begin{table}[htb]
    \centering
    \begin{tabular}{l|l|c|c}
        & &  {\em Lowercase Accuracy } & {\em Camel Case Accuracy} \\
         \hline
        \multirow{3}{*}{internal crawl} & baseline & 69.15\%  & 94.87\%  \\
        & pre-train & 80.89\%  & 91.49\%  \\
        & +fine-tune & {\bf 88.12\%}  & {\bf 96.20\%}  \\
         \hline
        \multirow{3}{*}{Common Crawl}& baseline & 63.64\%  & 85.48\%  \\
        & pre-train & 75.81\%  & 81.29\%  \\
        & +fine-tune & {\bf 85.21\%}  & {\bf 91.15\%}  \\         
    \end{tabular}
    \caption{Testing results on the internal crawl and Common Crawl data sets.}
    \label{tab:crawl_test}
\end{table}


\subsection{Error Analysis}
The pattern of improvement coming from pre-training is clear. Without pre-training, sometimes the model generates word-like segments. The pre-trained model is better at distinguishing real words and fakes ones. Table~\ref{tab:mixed_case_errors} shows some examples of success in the top region. 
What remains to be fixed are harder ones shown in the bottom region of Table~\ref{tab:mixed_case_errors}. 

\begin{table}[htb]
    \centering
    \begin{tabular}{l|l|l}
         & {\em Reference} & {\em Prediction} \\
         \hline
       \multirow{8}{*}{Errors fixed}  & \texttt{Mainspring Press} & \texttt{Mains pring Press} \\
        \multirow{8}{*}{{\em Prediction} $\rightarrow${\em Reference}} & \texttt{N M animal Control} & \texttt{N Manimal Control} \\
        & \texttt{artists ask art} & \texttt{artist saskart} \\
        & \texttt{planet earth} & \texttt{plane tearth} \\
        & \texttt{D C income}	& \texttt{D Cincome} \\
        & \texttt{Pubs history} & \texttt{Pubshistory} \\
        & \texttt{mens weekly} & \texttt{men sweekly} \\
        & \texttt{phoenix tennis} & \texttt{pho enix tennis} \\
         \hline
        \multirow{8}{*}{Errors uncorrected} & \texttt{Bulk S e o Tools} & \texttt{Bulk Seo Tools} \\
        \multirow{8}{*}{{\em Reference} $\rightarrow${\em Prediction}} & \texttt{Library U s gen net} & \texttt{Library Usgennet} \\
        & \texttt{just a film junkie} & \texttt{justa film junkie} \\
        & \texttt{pet lvr} & \texttt{pet l v r} \\
        & \texttt{ASAP Workouts} & \texttt{A S A P Workouts} \\
        & \texttt{S e o article} & \texttt{Seo article} \\
        & \texttt{iams company breeders} &\texttt{i a m s company breeders} \\
        & \texttt{u s a p a store} & \texttt{u s a pastore	}
    \end{tabular}
    \caption{Errors fixed and remaining uncorrected by pre-training.
    }
    \label{tab:mixed_case_errors}
\end{table}


\begin{figure}
    \centering
    \begin{tabular}{cc}
        \hspace{-0.5cm}\includegraphics[width=0.5\columnwidth]{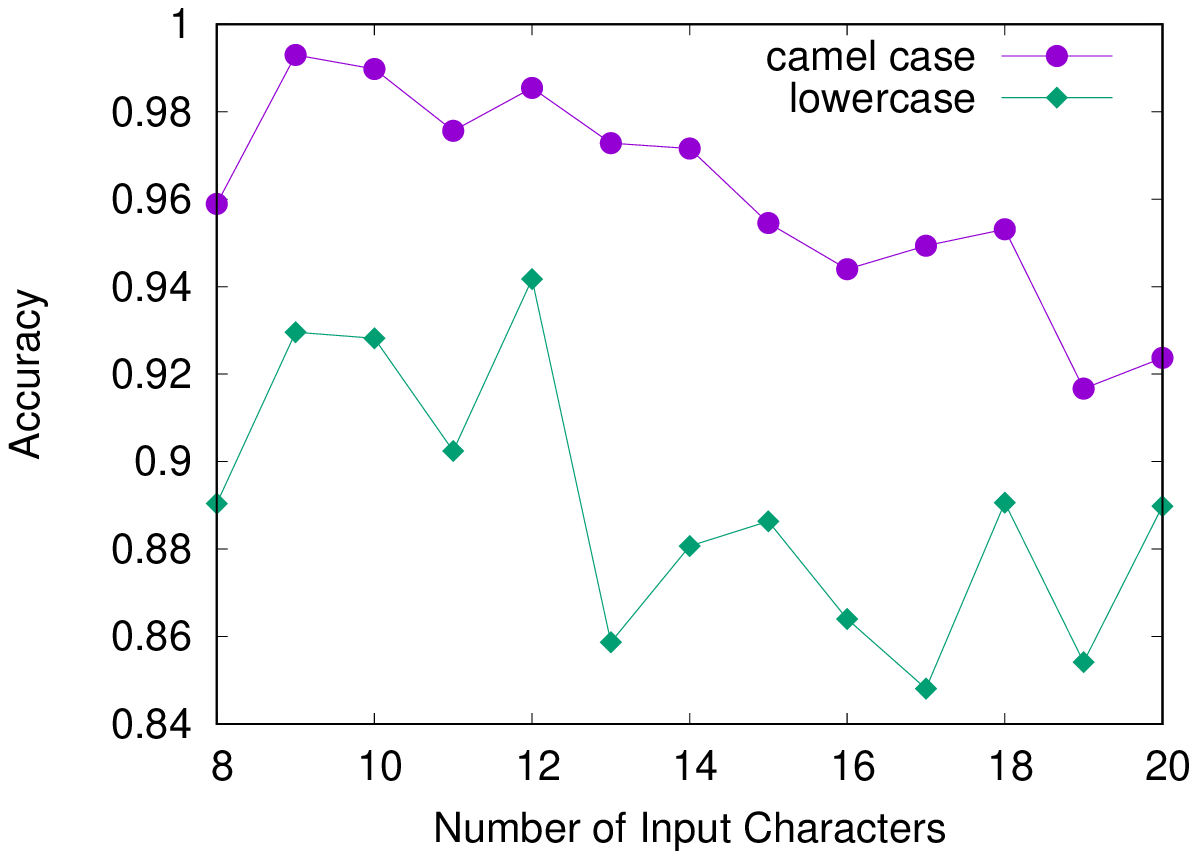} &                      \hspace{-0.5cm}\includegraphics[width=0.5\columnwidth]{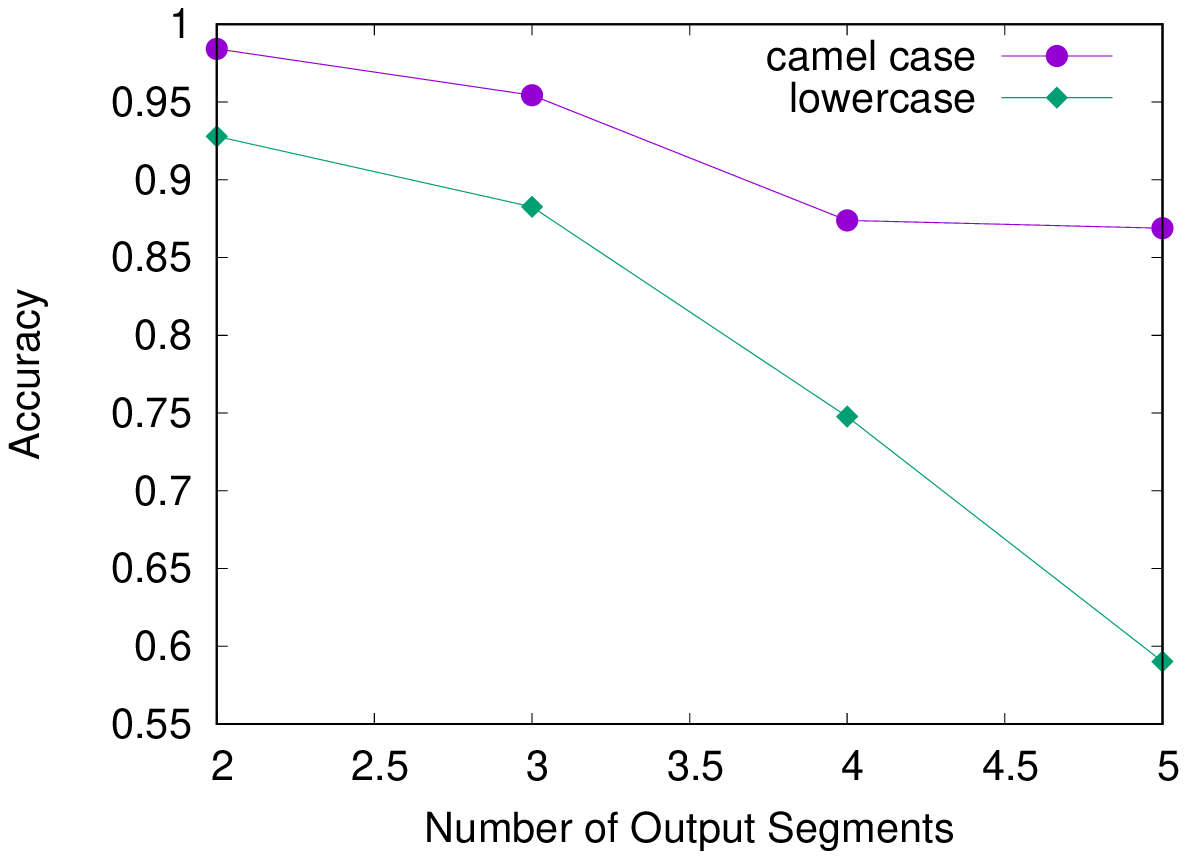} \\ \hspace{-0.5cm}\includegraphics[width=0.5\columnwidth]{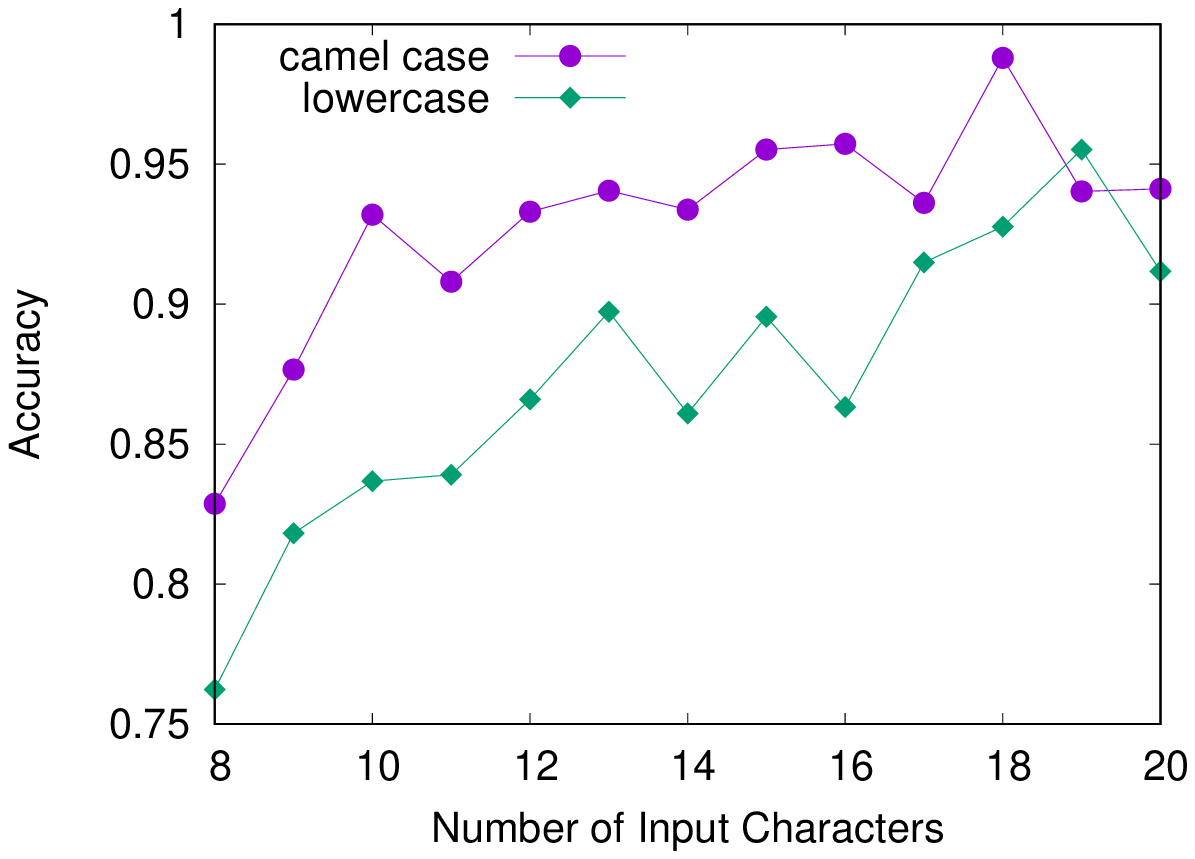} &
        \hspace{-0.5cm}\includegraphics[width=0.5\columnwidth]{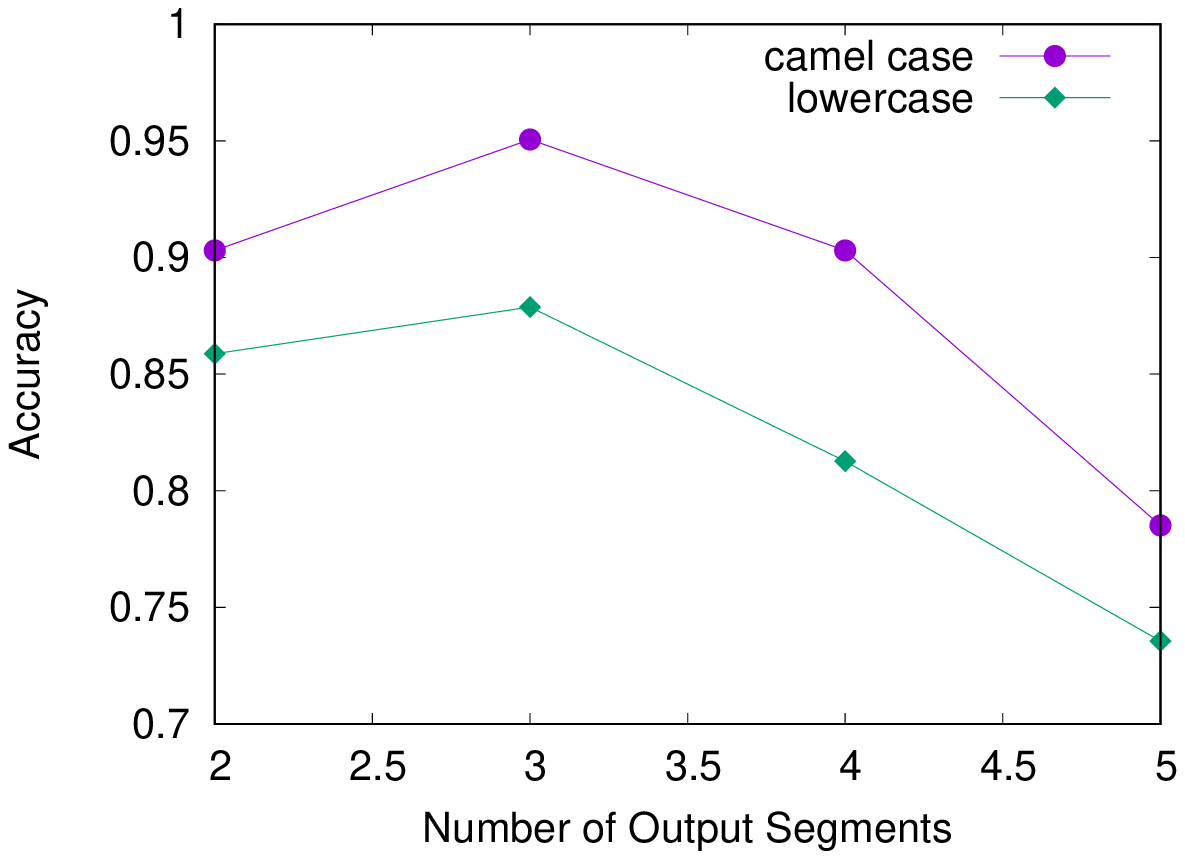} \\   
    \end{tabular}
    \caption{Accuracy as input/output length increases. The leftmost and rightmost lengths are bucketed. Top: internal crawl {\em dev}. Bottom: Common Crawl {\em dev}.}
    \label{fig:acc_len}
\end{figure}

Figure~\ref{fig:acc_len} shows how accuracy varies as the number of characters or segments contained in the input increases. The model is highly accurate on input with less than four segments. Due to more strict filtering in the internal crawl data set, there are very few examples with a single segment. There are many more single-segment examples in the Common Crawl data set. Single-segment examples are more challenging probably because they correspond to more rare words or terms. Prediction accuracy is not strongly correlated to input length as shown by the seemingly opposite trends on the two different data sets.

\label{sec:expts}

\section{Conclusion}
URL segmentation has applications in TTS and web search. Our contributions include a curated URL data set and a highly accurate RNN model boosted by pre-training on Knowledge Graph entities. 


\section*{Acknowledgments}
We thank Shankar Kumar, Corinna Cortes, and the reviewers for their comments and suggestions.
\newpage
\bibliography{anthology,acl2020}

\begin{thebibliography}{}

\bibitem[\protect\citename{Aharoni and
  Goldberg}2017]{aharoni-goldberg-2017-morphological}
Roee Aharoni and Yoav Goldberg.
\newblock 2017.
\newblock Morphological inflection generation with hard monotonic attention.
\newblock In {\em Proceedings of the 55th Annual Meeting of the Association for
  Computational Linguistics (Volume 1: Long Papers)}, pages 2004--2015,
  Vancouver, Canada, July. Association for Computational Linguistics.

\bibitem[\protect\citename{Bollacker \bgroup et al.\egroup
  }2008]{bollacker-etal-2008}
Kurt Bollacker, Colin Evans, Praveen Paritosh, Tim Sturge, and Jamie Taylor.
\newblock 2008.
\newblock Freebase: A collaboratively created graph database for structuring
  human knowledge.
\newblock In {\em Proceedings of the 2008 ACM SIGMOD International Conference
  on Management of Data}, SIGMOD '08, pages 1247--1250, New York, NY, USA. ACM.

\bibitem[\protect\citename{Chiang \bgroup et al.\egroup
  }2010]{chiang-etal-2010-bayesian}
David Chiang, Jonathan Graehl, Kevin Knight, Adam Pauls, and Sujith Ravi.
\newblock 2010.
\newblock {B}ayesian inference for finite-state transducers.
\newblock In {\em Human Language Technologies: The 2010 Annual Conference of
  the North {A}merican Chapter of the Association for Computational
  Linguistics}, pages 447--455, Los Angeles, California, June. Association for
  Computational Linguistics.

\bibitem[\protect\citename{Devlin \bgroup et al.\egroup
  }2019]{devlin-etal-2019-bert}
Jacob Devlin, Ming-Wei Chang, Kenton Lee, and Kristina Toutanova.
\newblock 2019.
\newblock {BERT}: Pre-training of deep bidirectional transformers for language
  understanding.
\newblock In {\em Proceedings of the 2019 Conference of the North {A}merican
  Chapter of the Association for Computational Linguistics: Human Language
  Technologies, Volume 1 (Long and Short Papers)}, pages 4171--4186,
  Minneapolis, Minnesota, June. Association for Computational Linguistics.

\bibitem[\protect\citename{Ebden and Sproat}2014]{ebden-sproat-2014}
Peter Ebden and Richard Sproat.
\newblock 2014.
\newblock The {K}estrel {TTS} text normalization system.
\newblock {\em Natural Language Engineering}, 21:333--353.

\bibitem[\protect\citename{Goldwater \bgroup et al.\egroup
  }2009]{goldwater-etal-2009-cognition}
Sharon Goldwater, Thomas Griffiths, and Mark Johnson.
\newblock 2009.
\newblock A bayesian framework for word segmentation: Exploring the effects of
  context.
\newblock {\em Cognition}, 112:21--54, 04.

\bibitem[\protect\citename{Hellwig and
  Nehrdich}2018]{hellwig-nehrdich-2018-sanskrit}
Oliver Hellwig and Sebastian Nehrdich.
\newblock 2018.
\newblock {S}anskrit word segmentation using character-level recurrent and
  convolutional neural networks.
\newblock In {\em Proceedings of the 2018 Conference on Empirical Methods in
  Natural Language Processing}, pages 2754--2763, Brussels, Belgium,
  October-November. Association for Computational Linguistics.

\bibitem[\protect\citename{Huang \bgroup et al.\egroup
  }2019]{huang-etal-2019-bert-segmentation}
Weipeng Huang, Xingyi Cheng, Kunlong Chen, Taifeng Wang, and Wei Chu.
\newblock 2019.
\newblock Toward fast and accurate neural chinese word segmentation with
  multi-criteria learning.
\newblock {\em CoRR}, abs/1903.04190.

\bibitem[\protect\citename{Ma \bgroup et al.\egroup }2016]{ma-etal-2016-letter}
Jianqiang Ma, Verena Henrich, and Erhard Hinrichs.
\newblock 2016.
\newblock Letter sequence labeling for compound splitting.
\newblock In {\em Proceedings of the 14th {SIGMORPHON} Workshop on
  Computational Research in Phonetics, Phonology, and Morphology}, pages
  76--81, Berlin, Germany, August. Association for Computational Linguistics.

\bibitem[\protect\citename{Ma \bgroup et al.\egroup }2018]{ma-etal-2018-state}
Ji~Ma, Kuzman Ganchev, and David Weiss.
\newblock 2018.
\newblock State-of-the-art {C}hinese word segmentation with bi-{LSTM}s.
\newblock In {\em Proceedings of the 2018 Conference on Empirical Methods in
  Natural Language Processing}, pages 4902--4908, Brussels, Belgium,
  October-November. Association for Computational Linguistics.

\bibitem[\protect\citename{Macherey \bgroup et al.\egroup
  }2011]{macherey-etal-2011-language}
Klaus Macherey, Andrew Dai, David Talbot, Ashok Popat, and Franz Och.
\newblock 2011.
\newblock Language-independent compound splitting with morphological
  operations.
\newblock In {\em Proceedings of the 49th Annual Meeting of the Association for
  Computational Linguistics: Human Language Technologies}, pages 1395--1404,
  Portland, Oregon, USA, June. Association for Computational Linguistics.

\bibitem[\protect\citename{Olivier}1968]{Olivier:68}
D.~C. Olivier.
\newblock 1968.
\newblock {\em Stochastic Grammars and Language Acquisition Mechanisms}.
\newblock {Ph.D.} thesis, Harvard University, Cambridge, MA.

\bibitem[\protect\citename{Peng \bgroup et al.\egroup
  }2004]{peng-etal-2004-chinese}
Fuchun Peng, Fangfang Feng, and Andrew McCallum.
\newblock 2004.
\newblock {C}hinese segmentation and new word detection using conditional
  random fields.
\newblock In {\em {COLING} 2004: Proceedings of the 20th International
  Conference on Computational Linguistics}, pages 562--568, Geneva,
  Switzerland, aug 23{--}aug 27. COLING.

\bibitem[\protect\citename{Peters \bgroup et al.\egroup
  }2017]{peters-etal-2017-semi}
Matthew Peters, Waleed Ammar, Chandra Bhagavatula, and Russell Power.
\newblock 2017.
\newblock Semi-supervised sequence tagging with bidirectional language models.
\newblock In {\em Proceedings of the 55th Annual Meeting of the Association for
  Computational Linguistics (Volume 1: Long Papers)}, pages 1756--1765,
  Vancouver, Canada, July. Association for Computational Linguistics.

\bibitem[\protect\citename{Sotelo \bgroup et al.\egroup
  }2017]{sotelo-etal-2017}
Jose Sotelo, Soroush Mehri, Kundan Kumar, João~Felipe Santos, Kyle Kastner,
  Aaron Courville, and Yoshua Bengio.
\newblock 2017.
\newblock Char2wav: End-to-end speech synthesis.
\newblock In {\em Proceedings of the 5th International Conference on Learning
  Representations, Workshop Track}, Toulon, France, April.

\bibitem[\protect\citename{Sproat \bgroup et al.\egroup
  }2001]{Sproat:EtAl:2001}
Richard Sproat, Alan Black, Stanley Chen, Shankar Kumar, Mari Ostendorf, and
  Christopher Richards.
\newblock 2001.
\newblock Normalization of non-standard words.
\newblock {\em Computer Speech and Language}, 15(3):287--333.

\bibitem[\protect\citename{Srinivasan \bgroup et al.\egroup
  }2012]{srinivasan-etal-2012-hastags}
Sriram Srinivasan, Sourangshu Bhattacharya, and Rudrasis Chakraborty.
\newblock 2012.
\newblock Segmenting web-domains and hashtags using length specific models.
\newblock In {\em Proceedings of the 21st ACM International Conference on
  Information and Knowledge Management}, CIKM '12, pages 1113--1122, New York,
  NY, USA. ACM.

\bibitem[\protect\citename{Wang \bgroup et al.\egroup
  }2011]{wang-etal-2011-webnlp}
Kuansan Wang, Christopher Thrasher, and Bo-June~Paul Hsu.
\newblock 2011.
\newblock Web scale nlp: A case study on url word breaking.
\newblock In {\em Proceedings of the 20th International Conference on World
  Wide Web}, WWW '11, pages 357--366, New York, NY, USA. ACM.

\bibitem[\protect\citename{Wang \bgroup et al.\egroup
  }2017]{Wang-etal-2017-tacotron}
Yuxuan Wang, RJ~Skerry-Ryan, Daisy Stanton, Yonghui Wu, Ron~J. Weiss, Navdeep
  Jaitly, Zongheng Yang, Ying Xiao, Zhifeng Chen, Samy Bengio, Quoc Le, Yannis
  Agiomyrgiannakis, Rob Clark, and Rif~A. Saurous.
\newblock 2017.
\newblock Tacotron: Towards end-to-end speech synthesis.
\newblock {\em Computing Research Repository}, arXiv:1703.10135.
\newblock version 2.

\bibitem[\protect\citename{Zhang and Clark}2007]{zhang-clark-2007-chinese}
Yue Zhang and Stephen Clark.
\newblock 2007.
\newblock {C}hinese segmentation with a word-based perceptron algorithm.
\newblock In {\em Proceedings of the 45th Annual Meeting of the Association of
  Computational Linguistics}, pages 840--847, Prague, Czech Republic, June.
  Association for Computational Linguistics.

\end{thebibliography}
\bibliographystyle{coling}

\end{document}